\pdfoutput=1

\documentclass[11pt]{article}

\usepackage{acl}

\usepackage{times}
\usepackage{latexsym}

\usepackage[T1]{fontenc}

\usepackage[utf8]{inputenc}

\usepackage{microtype}

\usepackage{subfig}
\usepackage{tabularx}
\usepackage{multirow}
\usepackage{booktabs}
\usepackage[most]{tcolorbox}
\newtcolorbox{mytextbox}[1][]{%
    sharp corners,
    enhanced,
    colback=white,
    top=5pt,
    bottom=5pt,
    left=8pt,
    right=8pt,
    height=1.5cm,
    width=6.5cm,
    attach title to upper,
    #1
}

\makeatletter
\def\@fnsymbol#1{\ensuremath{\ifcase#1\or \dagger\or \ddagger\or
   \mathsection\or \mathparagraph\or \|\or **\or \dagger\dagger
   \or \ddagger\ddagger \else\@ctrerr\fi}}
\makeatother

\usepackage{xcolor}

\title{Testing Pre-trained Language Models' Understanding of Distributivity via Causal Mediation Analysis}

\author{Pangbo Ban\Thanks{ Equal contribution.},
        Yifan Jiang\footnotemark[1],
        Tianran Liu\footnotemark[1],
        Shane Steinert-Threlkeld \\ 
        Department of Linguistics, University of Washington \\
        \texttt{\{pbban, yfjiang, tianranl, shanest\}@uw.edu}}

\begin{document}
\maketitle

\begin{abstract}
To what extent do pre-trained language models grasp semantic knowledge regarding the phenomenon of distributivity? In this paper, we introduce DistNLI, a new diagnostic dataset for natural language inference that targets the semantic difference arising from distributivity, and employ the causal mediation analysis framework to quantify the model behavior and explore the underlying mechanism in this semantically-related task. We find that the extent of models' understanding is associated with model size and vocabulary size. We also provide insights into how models encode such high-level semantic knowledge. Our dataset and code are available on \href{https://github.com/1fanj/CMA-distributivity}{GitHub}.
\end{abstract}

\section{Introduction}

The ability to understand and utilize semantic knowledge (consciously or unconsciously) is essential to human reasoning process. Although significant progress has been made by large-scale pre-trained language models on many reasoning-required tasks, it is still unclear whether these models have reached a considerable level of competence in discerning and processing semantic knowledge. To break into the black box, recent studies employ various analysis methods and bring evidence that semantic knowledge \citep{bowman-etal-2015-recursive, ettinger-2020-bert, jumelet-etal-2021-language} is encoded by pre-trained models. However, some issues still remain. First, due to the difficulty in being analyzed and probed for, many semantic phenomena are not touched on by NLP researchers, even though they have been studied by linguists for decades. Second, current analysis methods are not flawless. For example, \citet{Belinkov2021ProbingCP} reviews some limitations of the probing classifier paradigm such as the spurious correlation between the probing classifier and the original model.

To address these issues, in this paper, we leverage causal mediation analysis (CMA), a new analysis framework introduced by \citet{Vig2020CausalMA, Vig2020InvestigatingGB}, to test pre-trained language models' understanding of semantic knowledge, with a specific focus on predicative distributivity. As a complex linguistic phenomenon, predicative distributivity involves semantics, pragmatics, and psycholinguistics. With minimal pairs differing in distributivity, we look into whether pre-trained language models grasp the semantic difference, and how much a model component plays a role in such extent of understanding. Our contributions are as follows: 
\begin{itemize}
    \item 
    We introduce DistNLI, a diagnostic NLI dataset which targets testing pre-trained language models' ability of discerning the property of predicative distributivity via minimal pairs of coordinated sentences (Section~\ref{sec:distnli}).
    \item 
    We refine metrics used in the CMA framework, namely total effect (TE), natural indirect effect (NIE) and natural direct effect (NDE) to guarantee the effect decomposition. We apply the framework to the ternary NLI task (Sections~\ref{sec:exp-te} and \ref{sec:exp-nie}).
    \item 
    We find that pre-trained models with either more parameters or richer vocabulary show some understanding of distributivity. We also find that knowledge of distributivity is concentrated in middle layers and the level of concentration is patterned with the degree of understanding (Sections~\ref{sec:models}, \ref{sec:result-te} and \ref{sec:result-nie}).
\end{itemize}

\section{Related Work}

\paragraph{Distributivity}
A sentence with the verb predication applying to a group, e.g.\ \emph{Sumon and Frank built a boat.}, can be interpreted into at least two readings: the distributive and the collective ones \citep{scha1981collective}. In the distributive reading, the predication applies to each individual in the group (e.g.\ both Sumon and Frank individually/separately built a boat), whereas in the collective reading, the predication only applies to the group as a whole (e.g.\ Sumon and Frank built a boat \emph{jointly}).

To theorize this linguistic phenomenon, \citet{scha1981collective} introduces the property of distributivity as a tool to formalize the collective and distributive senses of a predication. While Scha analyzes distributivity as a pure lexical property of different predicates, \citet{link2002logical} propose that distributivity is a semantic operator comparable to \textit{each}, which he defines as the `D-operator'. Some later semanticists settle their analyses on the middle ground where the D-operator and the direct predication (based solely on the lexical property) theory are useful under different circumstances \cite{dowty1987collective,roberts1987modal,hoeksema1988semantics,verkuyl1993distributivity,winter1997choice,de2017two,champollion2017parts}. Arguing that the predication of a simple sentence can be directly interpreted based on its lexical meaning while the analysis of a complex one will need a D-operator, \citet{winter1997choice} further defines P-distributivity and Q-distributivity, which correspond to the distributive sense under the direct predication and the D-operator respectively. For instance, \textit{Azul and Marsha hold three balloons} falls in the category of Q-distributivity, since it is necessary to introduce the D-operator to analyze the distributivity given the group \textit{Azul and Marsha}. The example \textit{Yu and Vivian laughed}, on the other hand, is a case of P-distributivity, since it is apparent that the predicate \textit{laughed} entails the distributivity originated from its lexical meaning. 

Adopting the terminology proposed by \citet{de2017two}, we investigate the predicative distributivity in this paper by following the approach advocated by \citet{winter1997choice} and \citet{champollion2017parts} that collectivity and P-distributivity are categorized under the direct predication, which is paralleled to the D-operator that gives the Q-distributivity. We will call predicates which can be perceived in both distributive and collective senses ambiguous predicates in this paper, which corresponds to the mixed predicates in the previous literature. \cite{de2017two,champollion2017parts}.

\paragraph{Natural Language Inference}
Natural Language Inference (NLI) is the task of determining whether one sentence (premise) entails, contradicts, or is neutral to another sentence (hypothesis). Early attempts include \newcite{chen-etal-2017-enhanced} and \newcite{ghaeini-etal-2018-dr} which are LSTM-based. Recent progress in pre-trained language models such as BERT \cite{devlin-etal-2019-bert}, RoBERTa \cite{liu2019roberta}, and DeBERTa \cite{he2021deberta} and NLI datasets like SNLI \cite{bowman-etal-2015-large}, XNLI \cite{conneau-etal-2018-xnli}, and MNLI \cite{williams-etal-2018-broad} provides more opportunities to tackle this task. Pre-trained models fine-tuned on NLI datasets often achieve satisfying performance.  

However, diagnostic studies show that high scores achieved by neural models do not mean they truly understand the relationship between sentences. For example, \newcite{mccoy-etal-2019-right} find BERT trained on MNLI instead leverages shallow heuristics to make predictions. 

\paragraph{Causal Mediation Analysis}
Recently, several analysis methods have been proposed to reveal information that are learned and utilized by language models. One of them is causal mediation analysis \cite{robins1992identifiability, Pearl2001DirectAI}, a statistical framework to identify direct and indirect effects of an intervention on an outcome of interest. It is first introduced as an analysis method to the NLP field by \citet{Vig2020CausalMA, Vig2020InvestigatingGB} to scrutinize gender bias in language models. More recent studies use it to explore linguistic phenomena such as syntactic agreement \cite{finlayson-etal-2021-causal} and negation \cite{dobreva-keller-2021-investigating}. The framework consists of three metrics: TE, NIE and NDE. TE is used to quantify how an \textit{input intervention} (e.g., text edits) would affect a \textit{response variable} (e.g., predicted probabilities). NIE and NDE are used to measure the mediated influence realized through an intermediate variable, or \textit{mediator}, which can be a neuron, a whole layer, or an attention head. TE is usually decomposed into the sum of NIE and NDE. As \citet{Vig2020CausalMA, Vig2020InvestigatingGB} suggest, the CMA framework has great potential for extensions. Motivated by their work, in this paper, we apply the framework, which was limited to binary classification, to the NLI task and propose alternative definitions of the metrics for a more robust effect decomposition.

\section{The DistNLI Dataset}\label{sec:distnli}

\begin{table*}[ht]
\centering
\begin{tabularx}{0.98\textwidth}{XXll}
    \toprule
    \textbf{Premise} & \textbf{Hypothesis} & \textbf{NLI Label} & \textbf{Distributivity} \\
    \midrule
    Mia and Lin \textit{laughed}. & Mia \textit{laughed}./Lin \textit{laughed}. & Entailment & Distributive\\
    \multirow{2}{=}{Mia and Lin \textit{pushed a rock}.} & \multirow{2}{=}{Mia \textit{pushed a rock}./Lin \textit{pushed a rock}.} & Entailment & Distributive\\
    & & non-Entailment & Collective\\
    Mia and Lin \textit{gathered}. & $^\ast$Mia \textit{gathered}./$^\ast$Lin \textit{gathered}. & N/A & Collective\\
    \bottomrule
\end{tabularx}
\caption{\label{tab:nli_eg} The relationship between distributivity and NLI labels illustrated by examples from DistNLI. The NLI label is determined by the type of the predicate in the premise, and \textit{non-entailment} means that both \textit{neutral} and \textit{contradiction} are acceptable as true labels. Here, \textit{pushed a rock} is an ambiguous predicate, so it can be assigned with both NLI labels. }
\end{table*}

\subsection{Capturing distributivity with NLI} 

Although semanticists differ in the treatment of lexical distributivity (i.e., P-distributivity) in their theories, the evaluation of predicative distributivity generally relies on the validity of the inference from a predication of a group and the part with the \textit{each} operator in the case of conjunction (e.g.\ \emph{Sumon and Frank each built a boat}) \cite{dowty1987collective,lasersohn1995plurality,winter1997choice,champollion2017parts,de2017two}. 

Rooted in this definition, distributive predicates sanction the entailment relation between the plurality and its part, whereas collective or ambiguous predicates do not. Consequently, we use the NLI task to evaluate models' understanding of predicative distributivity. In our approach, the model's ability to discern distributivity is evaluated by the divergent prediction of models between predicates with differing 
distributivity. Table~\ref{tab:nli_eg} demonstrates the relationship between distributivity and NLI labels. For instance, given the premise \emph{Mia and Lin laughed} and the hypothesis \emph{Lin laughed}, the model should predict the label as entailment based on the distributive predicate \emph{laughed}. On the other hand, given the premise \emph{Mia and Lin pushed a rock} and the hypothesis \emph{Mia pushed a rock}, since \emph{pushed a rock} is an ambiguous predicate, the model will exhibit a completely different performance if it grasps the disparity in semantics the distributivity exerts.

\subsection{Data Generation}
We generate a synthetic NLI dataset consisting of premise-hypothesis pairs with \textbf{\texttt{[DP1] and [DP2] [Pred]}} as premise and\textbf{\texttt{[DP1]/[DP2] [Pred]}} as hypothesis. \textbf{\texttt{[DP1]}} and \textbf{\texttt{[DP2]}} denote determiner phrases and \textbf{\texttt{[Pred]}} denotes a predicate. For example:

\begin{center}
\begin{mytextbox}
    \textbf{Premise}: Mia and Lin wore a mask.
    \textbf{Hypothesis}: Mia (Lin) wore a mask.
\end{mytextbox}
\end{center}

None of determiner phrase contains quantifiers in its structure, and both lexical and phrasal predicates are included \cite{champollion2017parts}. Three kinds of noun phrases have been formulated, namely person, animal, and object, and it is guaranteed that no group nouns like \textit{the committee} and conventionalized conjunctions like \textit{Simon and Garfunkel} are included. The template is further instantiated with distributive and ambiguous predicates. We scrape existing categorized predicates from past publications on distributivity, and augment the list with predicates of similar pattern and characteristics regarding the semantic ambiguity in the information structure \cite{kroch1974semantics,taub1989collective,de2017two,champollion2017parts,coppock2019invitation}. The augmented list aligns with the report of \citet{safir1987binominal} that most predications with indefinite cardinal as the determiner manifest the distributive feature. 

\subsection{Annotation}

The annotation on predicative distributivity in this dataset consists of two stages. During the first stage, we recruit three graduate students, who are native speakers of American English with both linguistics and NLP background, to annotate grammatical sentences with predicates for whether they are distributive, collective or ambiguous. An example with pictures and explanation is given prior to the task. Considering the subtle nature of distributivity, we synthesize their judgements and discard highly controversial data points (i.e.\ predicates which received three distinct labels). During the second stage, the dataset is further confirmed by an expert in both Semantics and NLP to validate the result and guarantee a trustworthy dataset. The post-annotated data is split into the control group and the intervention group, with the former containing 164 pairs with distributive predicates and the latter containing 164 pairs with ambiguous predicates. The construction of the groups is explained in Section~\ref{sec:exp-te}. They together form the final DistNLI dataset of 328 premise-hypothesis pairs.

\section{Models}\label{sec:models}

\begin{table*}[ht]
\centering
\begin{tabular}{lcccccc}
    \toprule
    & \multicolumn{2}{c}{\textbf{ConjNLI}} & \multicolumn{2}{c}{\textbf{HANS}} & \multicolumn{2}{c}{\textbf{DistNLI}}\\
    \textbf{Model} & \textbf{Total Acc} & \textbf{AND Acc} & \textbf{Ent. Acc} & \textbf{Non-Ent. Acc} & \textbf{Dis. Acc} & \textbf{Amb. Acc}\\
    \midrule
    DeBERTa-base & 65.81 & 64.84 & 99.3 & 53.25 & 100.00 & 0.00\\
    DeBERTa-large & 66.61 & 65.99 & 99.89 & 54.81 & 100.00 & 0.00 \\
    DeBERTa-xlarge & 65.01 & 64.84 & 100 & 42.33 & 100.00 & 0.00\\
    DeBERTa-v2-xlarge & 66.29 & 65.99 & 99.96 & 49.87 & 100.00 & 0.00\\
    DeBERTa-v2-xxlarge & 66.45 & 65.42 & 99.92 & 43.35 & 100.00 & 0.00\\
    RoBERTa-large & 64.53 & 64.55 & 99.65 & 46.61 & 100.00 & 0.61\\
    \bottomrule
\end{tabular}
\caption{\label{tab:preexam} Pre-examination results on ConjNLI, HANS and DistNLI. For ConjNLI, \textbf{Total Acc} is the accuracy on the full ConjNLI data, and \textbf{AND Acc} is the accuracy on cases with \textit{and} in the premise, hypothesis, or both. 
For HANS, \textbf{Ent. Acc} and \textbf{Non-Ent. Acc} stand for \textbf{Entailed Acc} and \textbf{Non-Entailed Acc}, which are the accuracy on entailed cases and non-entailed cases respectively. For DistNLI, \textbf{Dis. Acc} and \textbf{Amb. Acc} stand for \textbf{Distributive Acc} and \textbf{Ambiguous Acc}, which are the accuracy on the control group and intervention group respectively. }
\end{table*}

\subsection{Model Selection}

We choose recent pre-trained models that are finetuned on MNLI as our target models. We try to control conditions as much as possible, such as model size and training setup. We use six models: DeBERTa (base, large, xlarge), DeBERTa-v2 (xlarge, xxlarge), and RoBERTa-large. RoBERTa-large shares the same vocabulary of size 50K as DeBERTa variants while DeBERTa-v2 variants increase their vocabulary size to 128K.

\subsection{Pre-examination}

Pre-trained language models can leverage various types of information learned from the training corpus to tackle the downstream tasks. The one we want to test in this paper is distributivity, but there is other related information models may use to make predictions, such as coordination, which is used to generate our dataset, or lexical overlap between the hypothesis and premise. To minimize the effect of these confounders, we run selected models over existing diagnostic NLI datasets, namely ConjNLI \cite{saha-etal-2020-conjnli} and HANS \cite{mccoy-etal-2019-right}. In addition, we also report each model's accuracy on DistNLI as a preliminary evaluation of models' ability to recognize distributivity. Results on these datasets are shown in Table \ref{tab:preexam}.

\paragraph{ConjNLI}
ConjNLI is an NLI dataset testing both Boolean and non-Boolean usage of conjuncts including \textit{and}, \textit{or}, \textit{but}, and their combination with quantifiers and negation \cite{saha-etal-2020-conjnli}. We report each model's accuracy on the whole development set of ConjNLI as well as the subset where \textit{and} exists in the premise, the hypothesis, or both. We find that all selected models perform reasonably well on ConjNLI, suggesting that they can handle diverse Boolean and non-Boolean coordinated sentences to a certain extent.

\paragraph{HANS}
HANS is an NLI dataset testing whether models have adopted syntactic heuristics, e.g., the lexical overlap heuristic, the subsequence heuristic, the constituent heuristic during pre-training \cite{mccoy-etal-2019-right}. HANS is annotated with \textit{entailment} on which the heuristics make correct predictions, and \textit{non-entailment} (\textit{neutral} or \textit{contradiction}) on which the heuristics make incorrect predictions. Therefore, if a model can perform perfectly on the entailed cases but fails on the non-entailed cases, it may exploit the heuristics. For selected models, we report their accuracy on both entailed and non-entailed cases. As expected, we find that all models achieve nearly perfect performance on entailed cases. Nevertheless, they reach much higher scores on the non-entailed cases compared to the experiment with BERT by \newcite{mccoy-etal-2019-right}. This indicates that these models rely less on syntactic heuristics to solve the NLI task. 

\paragraph{DistNLI}
We report each model's accuracy on the control (distributive) group and the intervention (ambiguous) group. Because the control group is labeled with \textit{entailment} and the intervention group is labeled with \textit{non-entailment}, unsurprisingly results on DistNLI match the same pattern as that on HANS. However, the accuracy gap between the two groups is much bigger. Although this seems to suggest selected models heavily rely on syntactic heuristics and hence fail to recognize distributivity, results on ConjNLI and HANS show that selected models can handle at lease some of the non-Boolean coordinated sentences on which the heuristics provide no help. It is also worth noting that even for our recruited annotators, distributivity is a challenging phenomenon. Their responses demonstrate more variance in ambiguous predicates labeling. This implies accuracy may not be a good metric to evaluate models' understanding of distributivity since it does not account for the change of predicted probability when the decision is not flipped. Thus, we leverage the CMA framework for further investigation. 

\section{Total Effect}\label{sec:te}

\subsection{Experiment Design}\label{sec:exp-te}

\paragraph{Response Variable}
To quantify the model behavior, we follow the general idea proposed by \citet{finlayson-etal-2021-causal}. We define the response variable $y$ as the odds that a model (parameterized by $\theta$) predicts \textit{non-entailment} for a premise-hypothesis pair $S_I$, where $I$ is the set of possible readings of $S$ which may contain a distributive reading, a collective reading or both: 
\begin{equation}
    \begin{aligned}
    y(S_I) & = \text{Odds} (\textit{non-entailment} | S_I) \\
    & = \frac{P_{\theta}(\textit{non-entailment}|S_I)}{P_{\theta}(\textit{entailment}|S_I)}
    \end{aligned}
\end{equation}
The larger the $y(S_I)$, the more likely the model predicts \textit{non-entailment} on the input pair $S_I$. With this definition, we can transform the original question into a binary task: whether models have a stronger preference over \textit{non-entailment}  given a premise-hypothesis pair with a particular predicate.

We hypothesize that if a model has some understanding of distributivity, \textit{ceteris paribus}, an ambiguous pair should result in a larger predicted probability of \textit{non-entailment} than a distributive pair, even if the model predicts \textit{entailment} for both. In other words, $y(S_I)$ should be small when $I$ only contains a distributive reading but relatively large when $I$ contains both readings, provided that all other aspects of $S$ are equal.

\begin{figure}[t!]
    \centering
    \includegraphics[trim=150 70 160 80,clip,width=\columnwidth]{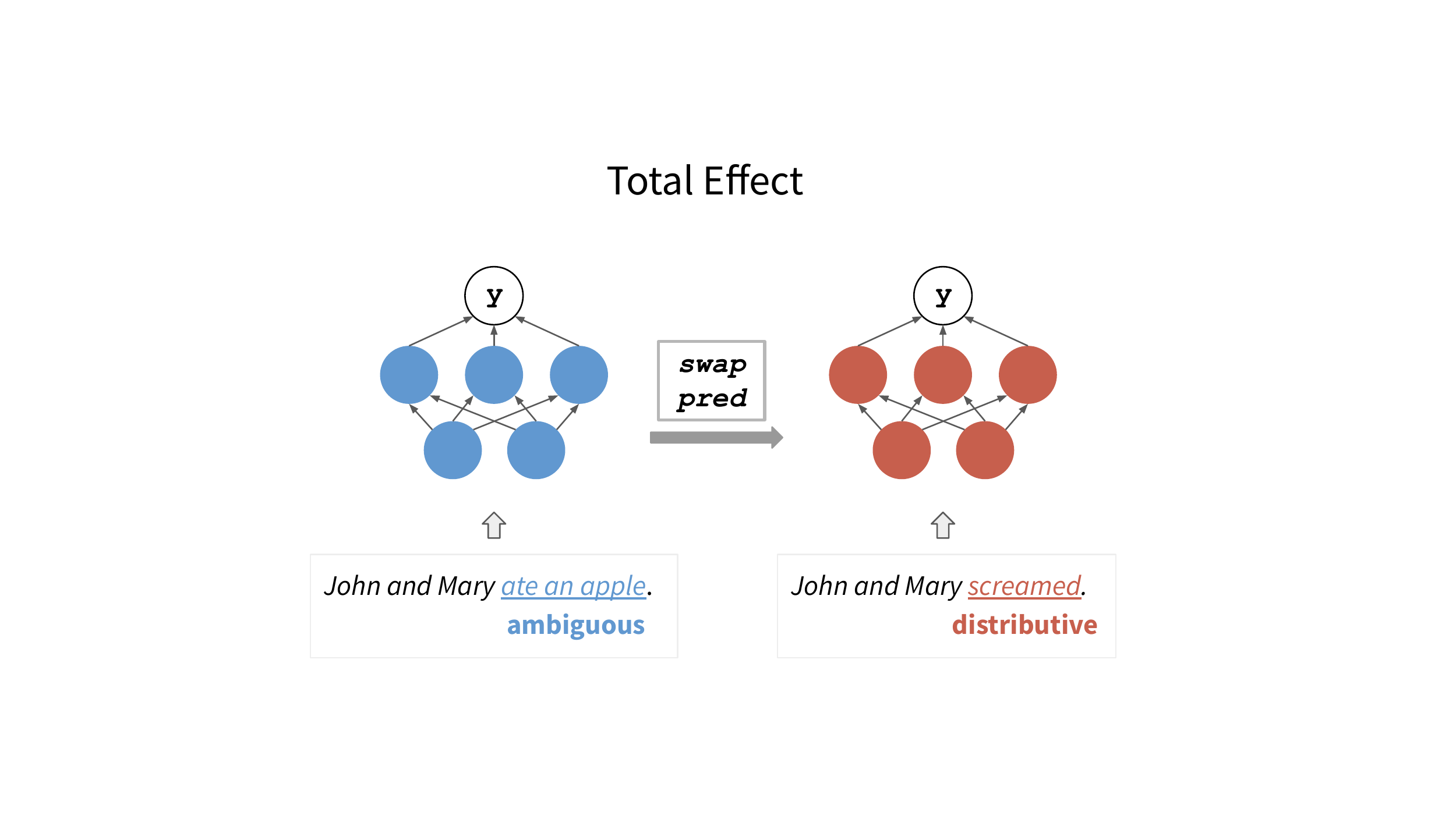}
    \caption{Total effect measures the relative change in $y$ from the input intervention which alters the distributivity of the predicate in the sentence.}
    \label{fig:cma-te}
\end{figure}

\paragraph{Input Intervention}
To isolate distributivity, we need a class of interventions to change possible readings of a given premise-hypothesis pair $S$ while keeping everything else the same. Since it is intractable to directly modify the possible readings $I$, we choose to modify the surface form. For our data templates, $I$ is determined by the predicate. Therefore, we define a do-operator \texttt{swap-pred}, which replaces the predicate in the given pair with a random sampled predicate of a different type, as illustrated in Figure \ref{fig:cma-te}.\footnote{The illustration of TE and the following NIE figure are inspired by \citet{Vig2020CausalMA,finlayson-etal-2021-causal}.} We also define the \texttt{null} operator which preserves the original predicate. The response variable $y(S_I)$ can be redefined as $y_i(S_p)$ where $i$ is a do-operator, $S$ is a premise-hypothesis pair, and $p$ indicates the type of the predicate in $S$:
\begin{alignat}{2}
    & y_{\texttt{null}}(S_p) = \text{Odds}(\textit{non-entailment} | S_p) \\
    & y_{\texttt{swap-pred}}(S_p) = \text{Odds}(\textit{non-entailment} | S_{p'})
\end{alignat}

Leveraging these interventions, we can split the input dataset into two groups: the control group and the intervention group. Pairs in the control group have distributive predicates with the \texttt{null} operator applied. Pairs in the intervention group are the same sentences but with the \texttt{swap-pred} operator applied. Each pair in the control group can have one or more matches in the intervention group. For example, if \textit{John and Mark smiled.} (premise) and \textit{John smiled.} (hypothesis) is in the control group, \textit{John and Mark built a house.} (premise) and \textit{John built a house.} (hypothesis) could be its potential match in the intervention group.

\paragraph{Metric}

TE is used to measure how much a response variable $y$ would change if we apply the \texttt{swap-pred} operator rather than the \texttt{null} operator. Instead of the odds difference definition used by previous studies \cite{Vig2020InvestigatingGB, finlayson-etal-2021-causal, dobreva-keller-2021-investigating, jeoung-diesner-2022-changed}, we adopt the odds ratio definition proposed by \citet{VanderWeele2010OddsRF}. To make the scale more symmetric, we take the equivalent logarithmic version:
\begin{equation}
    \begin{aligned}
        & \text{TE}(\texttt{swap-pred}, \texttt{null}; y, S_p) \\
        & = \log\left(\frac{y_{\texttt{swap-pred}}(S_p)}{y_{\texttt{null}}(S_p)}\right) \\
        & = \log\left(\frac{\text{Odds}(\textit{non-entailment} | S_{p'})}{\text{Odds}(\textit{non-entailment} | S_p)}\right)
    \end{aligned}
\end{equation}

We calculate the sample average total effect over DistNLI to estimate the average total effect over the population of all possible matched pairs:
\begin{equation}
    \begin{aligned}
        & \overline{\text{TE}}(\texttt{swap-pred}, \texttt{null}; y) \mathrel{\widehat{=}} \mu_{\text{TE}}
    \end{aligned}
\end{equation}

One benefit of this definition is that the total effect now inherits the interpretation of odds ratio, in addition to its own causal interpretation. Odds ratio is used to measure the strength of association between a response variable and an intervention. It compares the relative odds of the occurrence of an outcome of interest, given whether a particular intervention is performed \cite{Szumilas2010ExplainingOR}. Therefore, by analogy with the odds ratio, we can interpret $\overline{\text{TE}}$ in three cases: 
    (i) If $\overline{\text{TE}} > 0$, then the presence of ambiguous predicate in $S$ causes higher odds of \textit{non-entailment};
    (ii) If $\overline{\text{TE}} = 0$, then there is no causal relationship between the type of the predicate in $S$ and the model prediction;
    (iii)
    If $\overline{\text{TE}} < 0$, then the presence of ambiguous predicate in $S$ causes lower odds of \textit{non-entailment}.

Another benefit is that the lexical overlap heuristic is not a problem to our experiment, because $\overline{\text{TE}}$ measures the difference between sentences with swapped and unswapped predicates. If a model completely depends on the heuristic, $\overline{\text{TE}}$ will be close to 0 since the overlap between the premise and hypothesis remains the same when the input is intervened. In this case, no causal relationship is concluded between distributivity and the model prediction. If, however, we obtain a non-zero $\overline{\text{TE}}$, this should be due to factors other than the heuristic.

Since the size of DistNLI is relatively small, we perform the one-sample $t$-test with a significance level of 0.05 to infer about the average total effect over the full population. If $\overline{\text{TE}}$ is statistically significantly positive, we can conclude that the model has some understanding of distributivity.

\subsection{Result and Discussion}\label{sec:result-te}

\begin{table}[t!]
\centering
\begin{tabular}{lcccc}
    \toprule
    \textbf{Model} & \textbf{Mean} & \textbf{SD} & \textbf{T} & \textbf{P-value} \\
    \midrule
    D-b & 0.040 & 1.091 & 0.468 & 0.320 \\
    D-l & 0.314 & 0.900 & 4.452 & $<$7e-06 \\
    D-xl & 0.351 & 0.507 & 8.844 & $<$7e-16 \\
    D-v2-xl & 0.856 & 0.796 & 13.724 & $<$2e-29 \\
    D-v2-xxl & 0.828 & 1.088 & 9.724 & $<$3e-18\\
    R-l & 0.779 & 1.279 & 7.774 & $<$4e-13 \\
    \bottomrule
\end{tabular}
\caption{\label{tab:te} One Sample $t$-test of $\overline{\text{TE}}$ for each model. Here, R stands for RoBERTa, D stands for DeBERTa, b stands for base, and l stands for large.}
\end{table}

\begin{table}[t!]
    \includegraphics[width=\columnwidth]{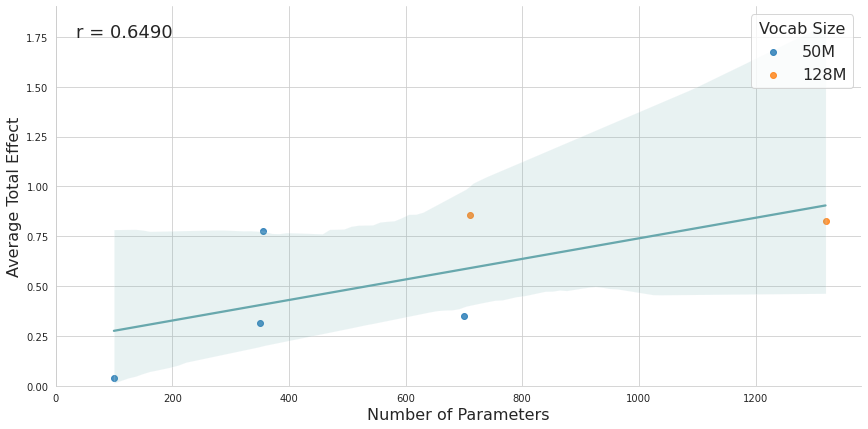}
    \captionof{figure}{\label{fig:te-params-vocabs}Relationship between $\overline{\text{TE}}$, the number of parameters and vocabulary size. Each point is a model.}
\end{table}

Table \ref{tab:te} presents a one-sample $t$-test of the average total effect for each model. Except for DeBERTa-base, all models have a significantly positive $\overline{\text{TE}}$ with a significant level of 0.05. Based on the interpretation of $\overline{\text{TE}}$, these models are able to discern distributivity to some extent.

\paragraph{Models with more parameters tend to show understanding of distributivity.} We find that $\overline{\text{TE}}$ is positively correlated with the number of parameters, as shown by Figure \ref{fig:te-params-vocabs} (\textit{r} = 0.649). While there are confounders such as vocabulary size, number of layers, pre-training task, etc., the trend holds when we control for model architecture and only consider DeBERTa variants. This finding may suggest that larger models have a stronger ability to capture linguistic phenomena presented in the training corpus. \citet{Vig2020CausalMA} report an analogous result on gender bias. We also observe that the effect on $\overline{\text{TE}}$ vanishes when the number of parameters increases: as shown by Table~\ref{tab:te}, DeBERTa-large has a $\overline{\text{TE}}$ eight times greater than DeBERTa-base, but merely 0.04 lesser than DeBERTa-xlarge. This observation is in line with the finding of \citet{K2020Cross-Lingual} that the number of parameters has little effect on model performance after a certain threshold.

\paragraph{Models with richer vocabulary tend to show understanding of distributivity.} We find that $\overline{\text{TE}}$ is associated with the size of vocabulary. As illustrated in Figure \ref{fig:te-params-vocabs}, DeBERTa-v2 variants have a $\overline{\text{TE}}$ of around 0.8, which is considerably larger than other models. We suspect this significant increase is due to their larger vocabulary size. Since distributivity is determined by the type of predicates, a richer vocabulary is expected to lead to a better semantic representation of predicates which in turn boosts $\overline{\text{TE}}$, although the effects of other confounders might not be ruled out. 

\section{Natural Indirect Effect}\label{sec:nie}

\subsection{Experiment Design}\label{sec:exp-nie}

\paragraph{Neuron Intervention}
To study the causal contribution of neurons, the hypothesized mediator in our experiment will be a single neuron or a group of neurons. Rather than investigating neurons for each input token independently \cite{Vig2020CausalMA, Vig2020InvestigatingGB, finlayson-etal-2021-causal}, we intervene on neurons for all input tokens simultaneously. This approach is computationally cheaper but still comprehensive enough to give a full picture of the underlying causal mechanism.\footnote{Pilot experiments focusing on the \texttt{[CLS]} token, an approach used by \citet{dobreva-keller-2021-investigating}, yielded mixed results, which is consistent with the findings of \citet{reimers-gurevych-2019-sentence} that the \texttt{[CLS]} token is not an ideal representation of sentence level meaning.}

\paragraph{Metric}

\begin{figure}[t!]
    \centering
    \includegraphics[trim=150 70 160 80,clip,width=\columnwidth]{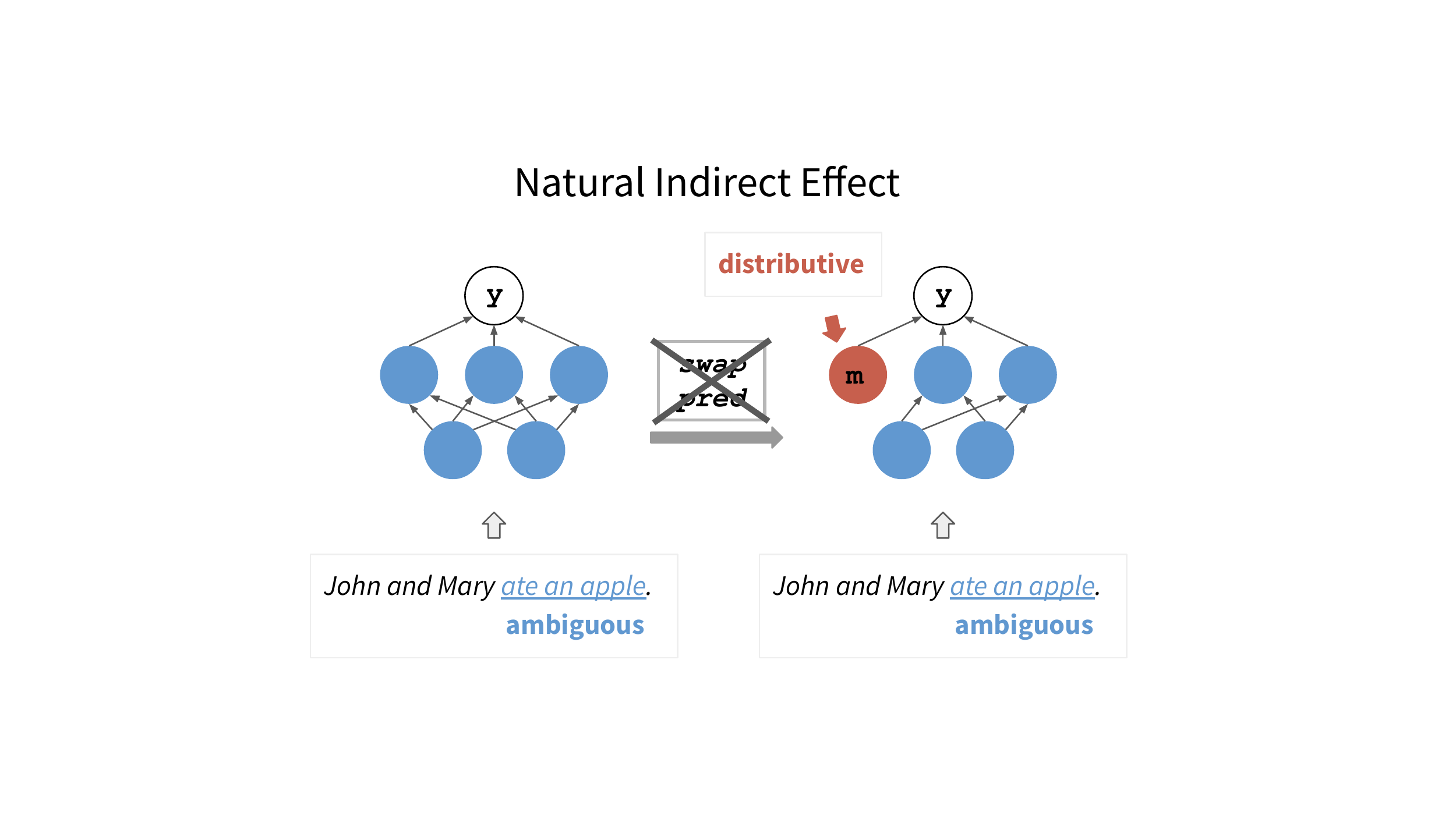}
    \caption{Natural indirect effect measures the relative change in $y$, given the presence of the input intervention, if every path into the mediator $m$ is blocked by setting $m$ to the value it would have been without the input intervention.}
    \label{fig:cma-nie}
\end{figure}

Natural Indirect Effect is used to measure how much the response variable $y$ would change with the \texttt{swap-pred} operator applied, if we set the hypothesized mediator $m$ to the value it would have been without rather than with the input intervention (demonstrated in Figure \ref{fig:cma-nie}). Similar to TE, we use the log odds ratio and estimate the population average natural indirect effect:
\begin{equation}
    \begin{aligned}
        & \text{NIE}(\texttt{swap-pred}, \texttt{null}; y, m, S_p) \\
        & = \log\left(\frac{y_{\texttt{swap-pred}}(S_p)}{y_{\texttt{swap-pred}, m_{\texttt{null}}}(S_p)}\right) \\
    \end{aligned}
\end{equation}
where $m$ is a hypothesized mediator and $m_{\texttt{null}}$ means that $m$ is set to the value it would have been in the absence of the input intervention. We can also define NDE in a similar way:
\begin{equation}
    \begin{aligned}
        & \text{NDE}(\texttt{swap-pred}, \texttt{null}; y, m, S_p) \\
        & = \log\left(\frac{y_{\texttt{swap-pred}, m_{\texttt{null}}}(S_p)}{y_{\texttt{null}}(S_p)}\right) \\
    \end{aligned}
\end{equation}
\citet{VanderWeele2010OddsRF} prove that the log odds ratio definition of causal effects holds a decomposition property: $\text{TE} = \text{NIE} + \text{NDE}$ even when there are interactions and nonlinearities.

The NIE and NDE defined above are in principle an implementation of what \citet{robins1992identifiability} refer to as ``total indirect effect'' and ``pure direct effect''. The NDE given by \citet{Vig2020CausalMA, Vig2020InvestigatingGB} follows the same idea, but their NIE instead formulates the "pure indirect effect". A consequence is that the decomposition property is only guaranteed for linear models \cite{Pearl2001DirectAI, Vig2020CausalMA}, which is a potential shortcoming as an analysis method for neural networks.

Given the decomposition property, $\overline{\text{NIE}}$ allows us to measure the magnitude of causal contribution a model component makes to the model behavior, which is quantified by $\overline{\text{TE}}$. In this respect, it potentially solves the problem of spurious correlation between the probing classifier and the original model \citep{Belinkov2021ProbingCP}. In our experiment, we use it to verify the causal relationship between the semantic information encoded in the original model and the prediction given by the NLI classifier. We can interpret the values of $\overline{\text{NIE}}$ similarly to  $\overline{\text{TE}}$.

\begin{figure*}[ht]
    \centering
    \subfloat[DeBERTa-large]{\includegraphics[width = \columnwidth]{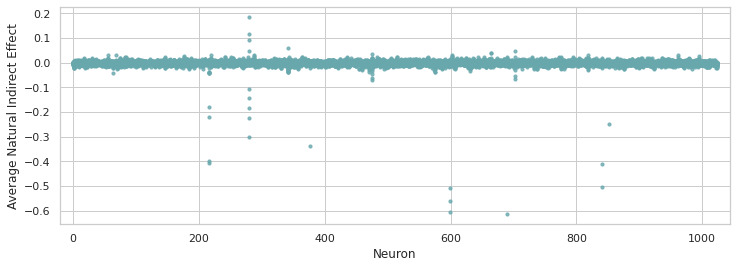}} 
    \subfloat[DeBERTa-xlarge]{\includegraphics[width = \columnwidth]{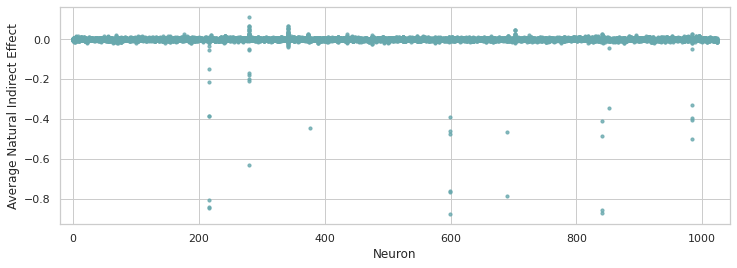}} \\
    \subfloat[DeBERTa-v2-xlarge]{\includegraphics[width = \columnwidth]{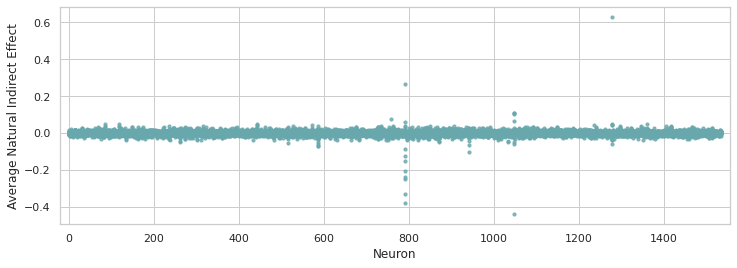}} 
    \subfloat[RoBERTa-large]{\includegraphics[width = \columnwidth]{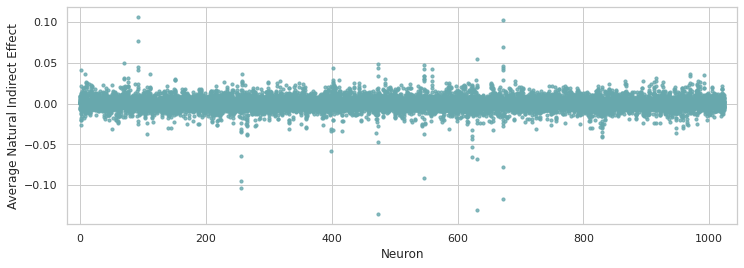}} 
    \caption{Neuron-wise $\overline{\text{NIE}}$ of the models that pass the $\overline{\text{TE}}$ threshold. The \textit{x}-axis represents the indices of neurons, which range from 0 to hidden size. The indices are not unique: neurons from different layers have the same index.}
    \label{fig:nie}
\end{figure*}

\subsection{Result and Discussion}\label{sec:result-nie}

\begin{figure}[t!]
    \centering
    \subfloat[]{\includegraphics[width = 0.25\columnwidth]{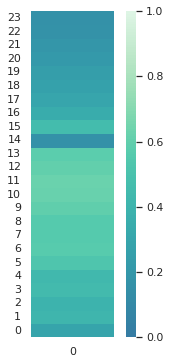}}
    \subfloat[]{\includegraphics[width = 0.25\columnwidth]{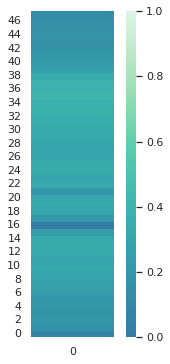}} 
    \subfloat[]{\includegraphics[width = 0.25\columnwidth]{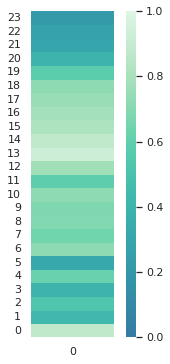}}
    \subfloat[]{\includegraphics[width = 0.25\columnwidth]{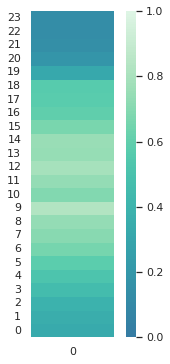}}
    \caption{Layer-wise $\overline{\text{NIE}}$ (top 1\% neurons) of the models that pass the $\overline{\text{TE}}$ threshold. From left to right are DeBERTa-large, DeBERTa-xlarge, DeBERTa-v2-xlarge and RoBERTa-large. The y-axis represents layers and the color represents values of the $\overline{\text{NIE}}$. }
    \label{fig:topknie}
\end{figure}

We experiment on models which pass the significance threshold.\footnote{Due to limited computational resources, DeBERTa-v2-xxlarge is excluded.} Figure \ref{fig:nie} illustrates the neuron-wise $\overline{\text{NIE}}$: for all models, most neurons have $\overline{\text{NIE}}$s around zero, but a few outliers can also be identified. In order to determine which neurons are responsible the most for the model behaviour, we also select the top $1\%$ of neurons with highest individual $\overline{\text{NIE}}$s from each layer and evaluate the layer-wise $\overline{\text{NIE}}$. Figure \ref{fig:topknie} illustrates the layer-wise $\overline{\text{NIE}}$ obtained from selected neurons.

\paragraph{Knowledge of distributivity is mostly concentrated in middle layers.} We define the depth of a layer as its  number divided by the total number of layers. Based on this metric, we can divide layers into three groups: early (0 - 0.33), middle (0.33 - 0.67), and final (0.67 - 1). In Figure~\ref{fig:topknie}, a concentration pattern of $\overline{\text{NIE}}$  is clearly shown by the color opacity of layers for all models. Specifically, middle layers have higher $\overline{\text{NIE}}$s than other layers. The exact layer where $\overline{\text{NIE}}$ peaks occur is more idiosyncratic, but still inside or near middle layers: 0.5 (layer \#11) for DeBERTa-large, 0.73 (layer \#35) for DeBERTa-xlarge, 0.54 (layer \#13) for DeBERTa-v2-xlarge, and 0.42 (layer \#9) for RoBERTa-large. This finding differs from the conclusion of \citet{tenney2018what} that semantic information is hardly localized in BERT-like models, although \citet{jawahar-etal-2019-bert} also report that most semantic tasks archive the best performance around middle layers.

\paragraph{Knowledge of distributivity is more concentrated in the models with higher degree of understanding.} We find that the level of concentration of $\overline{\text{NIE}}$ patterns with the magnitude of $\overline{\text{TE}}$: $\overline{\text{NIE}}$s are concentrated in fewer neurons in DeBERTa-v2-xlarge and RoBERTa-large (both have a $\overline{\text{TE}}$ of about 0.8) than in DeBERTa-large and DeBERTa-xlarge (both have a $\overline{\text{TE}}$ of about 0.3). This finding is supported by the following observations: First, as shown by Figure~\ref{fig:topknie}, top $1\%$ of neurons is sufficient to achieve the full total effect for the former two models, but not for the latter two models. Second, we notice that a few neurons have extremely higher $\overline{\text{NIE}}$ for the former two models. For example, neuron \#1279 at layer \#0 (located at the top right of Figure~\ref{fig:nie}c) in DeBERTa-v2-xlarge has a $\overline{\text{NIE}}$ of $0.6735$, much higher than most other neurons. According to the interpretation of $\overline{\text{NIE}}$, these neurons are causally and positively responsible for the model behaviour. The pattern is not observed in DeBERTa-large and DeBERTa-xlarge.

\section{Conclusion}

In this paper, we propose DistNLI, a diagnostic NLI dataset to examine to what extent pre-trained language models can discern the phenomenon of distributivity. By extending the CMA framework, we show that models including DeBERTa and RoBERTa have some understanding of distributivity, which provides further evidence that models have ability to encode high-level semantic knowledge, and reveal some interesting patterns related to the underlying mechanism of these models.

One direction for future improvement would be increasing the diversity of predicates and subjects in DistNLI. At present, we only look at subjects that are two coordinated DPs, but the phenomenon of distributivity applies to all noun phrases which denote groups and even without the utilization of conjuncts. It is possible that pre-trained language models can also differentiate more complicated combinations, as they are trained on large-scale text data. Another direction would be investigating how robust the CMA framework is to the definition of metrics, such as an empirical comparison between alternative definitions.

\section{Limitation}

Due to the specificity of the linguistic phenomenon involved and its size, this DistNLI dataset should only be used as a diagnostic dataset in the investigation of distributivity of verb predication. Also, occasionally some minimal pairs in the dataset could contradict with the world knowledge considering the nature of artificiality. On the one hand, the creators of this dataset have filtered out pairs that are tremendously deviant from the world knowledge by majority voting. On the other hand, even if there is still any deviating pair against the commonsense (i.e.\ \textit{The lion and the seal found a habitat}), the distributivity manifested in such examples will not be confounded as long as the grammaticality is guaranteed, since the extent of deviance is constant between the premise and the hypothesis.

\bibliography{anthology,custom}




\end{document}